# Topological Parameters for time-space tradeoff

**Rina Dechter**
Information and Computer Science
University of California, Irvine
*dechter@ics.uci.edu*

## Abstract

In this paper we propose a family of algorithms combining tree-clustering with conditioning that trade space for time. Such algorithms are useful for reasoning in probabilistic and deterministic networks as well as for accomplishing optimization tasks. By analyzing the problem structure it will be possible to select from a spectrum the algorithm that best meets a given time-space specification.

## 1   INTRODUCTION

Topology-based algorithms for constraint satisfaction and probabilistic reasoning fall into two distinct classes. One class is centered on tree-clustering, the other on cycle-cutset decomposition. Tree-clustering involves transforming the original problem into a tree-like problem that can then be solved by a specialized tree-solving algorithm [Mackworth and Freuder, 1985; Pearl, 1986]. The tree-clustering algorithm is time and space exponential in the induced width (also called *tree width*) of the problem's graph. The transforming algorithm identifies subproblems that together form a tree, and the solutions to the subproblems serve as the new values of variables in a tree metalevel problem. The metalevel problem is called a *join-tree*.

The cycle-cutset method exploits the problem's structure in a different way. A cycle-cutset is a subset of the nodes in a graph which cuts all of the graph's cycles. A typical cycle-cutset method enumerates the possible assignments to a set of cutset variables and, for each cutset assignment, solves (or reasons about) a tree-like problem in polynomial time. The overall time complexity is exponential in the size of the cycle-cutset [Dechter, 1992]. Fortunately, enumerating all the cutset's assignments can be accomplished in linear space.

Since the space complexity of tree-clustering can severely limit its usefulness, we investigate the extent to which its space complexity can be reduced, while reasonable time complexity guarantees are maintained. Is it possible to have the time guarantees of clustering while using linear space? On some problem instances, it is possible. Specifically, on those problems whose associated graph has a tree width and a cutset of comparable sizes (e.g., on a ring, the cutset size is 1 and the tree width is 2, leading to identical time bounds). In general, however, it is probably not possible, because in general the minimal cycle-cutset of a graph can be much larger than its tree width, namely, $r \leq c + 1$, where $c$ is the minimal cycle-cutset and $r$ is the tree width. Furthermore, we conjecture that any algorithm that has a time bound that is exponential in the tree-width will, on some problem instances, require exponential space in the tree width.

The space complexity of tree-clustering can be bounded more tightly using the *separator width*, which is defined as the size of the maximum subset of variables shared by adjacent subproblems in the join-tree. Our initial investigation employs separator width to control the time-space tradeoff. The idea is to combine adjacent subproblems joined by a large separator into one big cluster so that the remaining separators are of smaller size. Once a join-tree with smaller separators is generated, its potentially larger clusters can be solved using the cycle-cutset method.

We will develop such time-space tradeoffs for belief network processing, constraint processing, and optimization tasks, yielding a sequence of algorithms that can trade space for time. With this characterization it will be possible to select from a spectrum of algorithms the one that best meets some time-space requirement. Algorithm tree-clustering and cycle-cutset decomposition are two extremes in this spectrum.

We introduce the ideas using belief networks (section 2) and then show how they are applicable to constraint networks (section 3) and to optimization problems (section 4).

## 2   PROBABILISTIC NETWORKS

The observation that methods based on conditioning are inferior *time-wise* and superior space-wise in a worst-case sense to methods based on clustering is not widely acknowledged in belief network processing. This can be partially attributed to [Shachter *et al.*, 1991], where it is argued that conditioning is a special case of clustering. While there are always abstract criteria according to which one algorithm can be viewed a special case of another, one should not abstract away the hard fact that conditioning takes linear space while clustering is space hungry.

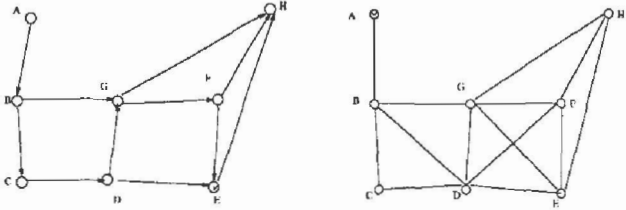

Figure 1: A belief network and its moral graph

A belief network is a concise description of a complete probability distribution. It is defined by a directed acyclic graph over nodes representing random variables, and each variable is annotated with the conditional probability matrices specifying its probability given each value combination of its parent variables. A belief network uses the concept of a directed graph.

**Definition 1** [Directed graph] A *directed graph* $G = \{V, E\}$, where $V = \{X_1, ..., X_n\}$ is a set of elements and $E = \{(X_i, X_j) | X_i, X_j \in V\}$ is the set of edges. If an arc $(X_i, X_j) \in E$, we say that $X_i$ points to $X_j$. For each variable $X_i$, $pa(X_i)$ is the set of variables pointing to $X_i$ in $G$, while $ch(X_i)$ is the set of variables that $X_i$ points to. The family of $X_i$ includes $X_i$ and its parent variables. A directed graph is acyclic if it has no directed cycles.

**Definition 2** [Belief Networks] Let $X = \{X_1, ..., X_n\}$ be a set of random variables over multi-valued domains, $D_1, ..., D_n$. A *belief network* is a pair $(G, P)$ where $G$ is a directed acyclic graph and $P = \{P_i\}$ are the conditional probability matrices over the families of $G$, $P_i = \{P(X_i | pa(X_i))\}$. An assignment $(X_1 = x_1, ..., X_n = x_n)$ can be abbreviated as $x = (x_1, ..., x_n)$. The belief network represents a probability distribution over $X$ having the product form

$$P(x_1, ...., x_n) = \Pi_{i=1}^n P(x_i | x_{pa(X_i)})$$

where $x_{pa(X_i)}$ denotes the projection of a tuple $x$ over $pa(X_i)$. An evidence set $e$ is an instantiated subset of variables. A *moral graph* of a belief network is an undirected graph generated by connecting any two head-to-head pointing arcs in the belief network's directed graph and removing the arrows.

Figure 1 shows a belief network's acyclic graph and its associated moral graph. Two of the most common tasks over belief networks are determining posterior beliefs and, given a set of observations finding the most probable explanation (MPE).

## 2.1  TREE-CLUSTERING

The most widely used method for processing belief networks is tree-clustering. Tree-clustering methods have two parts: determining the structure of the newly generated tree problem, and assembling the conditional probabilistic distributions between subproblems. The structure of the join-tree is generated using graph information only. First the moral graph is embedded in a chordal graph by adding some edges. This is normally accomplished by picking a variable ordering $d = X_1, ..., X_n$, then, moving from $X_n$ to $X_1$, recursively connecting all the neighbors of $X_i$ that precede it in the ordering. The *induced width* (or *tree width*) of this ordered graph, denoted $w*(d)$, is the maximal number of earlier neighbors in the resulting graph of each node. The maximal cliques in the newly generated chordal graph form a *clique-tree* and serve as the subproblems (or clusters) in the final tree, the join-tree. The induced width $w*(d)$ equals the maximal clique minus 1. The size of the smallest induced width over all the graph's clique-tree embeddings is the *induced width* (or, *tree-width*), $w*$ of the graph. A subset of nodes is called a *cycle-cutset* if their removal makes the graph cycle-free.

Once the tree structure is determined, each subproblem is viewd as a metavariable whose values are all the value combinations of the original variables in the cluster. The conditional probabilities between neighboring cliques can then be computed [Pearl, 1988]. Alternatively, the marginal probability distributions for each clique can be computed [Lauritzen and Spiegelhalter, 1988]. In both cases, the computation is exponential in the clique's size, so clustering is time and space exponential in the moral graph's induced-width [Pearl, 1988; Lauritzen and Spiegelhalter, 1988].

**Example 1:** Since the moral graph in Figure 1(b) is chordal no arc is added in the first step of clustering. The maximal cliques of the chordal graph are $\{(A, B), (B, C, D), (B, D, G), (G, D, E, F), (H, G, F, E)\}$, and an associated join-tree is given in Figure 2(a). The tree width of this problem is 3, the separator width is 3, and its minimal cycle-cutset has size 3 ($G, D, E$ is a cycle-cutset). A brute-force application of tree-clustering to this problem is time and space exponential in 4.

The implementation of tree-clustering can be tailored to particular queries, to yield better space complexity on some instances. For instance, one way to compute the belief of a particular variable, once the structure of a join-tree is available, is to generate a rooted join-tree whose root-clique contains the variable whose belief we wish to update. Then cliques can be processed recursively from leaves to the root. Processing a clique involves computing the marginals over the separators, which is defined as the intersection between the clique and its parent clique. A brute-force way of accomplishing this is to multiply the probabilistic functions in the clique for each clique's tuple and accumulate a sum over the variables in the clique that do not participate in the separator. Such computation is time exponential in the clique's size, but it requires only linear space to record the ouptput. In our case, the output space is exponential in the separator's size (See example in section 2.4).

Once the computation of a clique terminates, the computed marginal on its separator is added to the parent clique and the clique is removed. Computation resumes over new leaf-cliques until only the root-clique, containing the variable whose belief we wish to assess, remains[1]. Such a modification is applicable to various tasks and, in particular, to belief assessment and MPE.

This modification tightens the bound on space complexity by using the separator width of the join-tree. The separator width of a join-tree is the maximal size of the intersections between any two cliques, and the separator width of a graph is the minimal separator width among the separator widths over all the graph's clique-tree embeddings.

In summary,

**Theorem 1:** *[time-space of clustering]* *Given a belief network whose moral graph can be embedded in a clique-tree having induced width $r$ and separator width $s$, the time complexity for determining the beliefs and the MPE is $O(n \cdot exp(r))$ while the space complexity is $O(n \cdot exp(s))$.* □

Clearly $s \leq r$. Note that since in our example the separator width is 3 and the tree width is also 3, we do not gain much space-wise by the modified algorithm outlined above. There are, however, many cases where the separator width is much smaller than the tree width.

## 2.2   CUTSET CONDITIONING

Alternatively, belief networks may be processed by cutset conditioning [Pearl, 1988]. Conditioning computes conditioned beliefs and MPE for each assignment to a cycle-cutset, using a tree algorithms applied to the tree resulting from deleting the conditioning variables, and then computes the overall belief by taking a weighted sum or performing a maximization. The weights are the belief associated with each cutset assignment. The time complexity of conditioning is therefore worst-case exponential in the cycle-cutset size of the network's moral graph and it is space linear. This latter fact is not obvious since it requires computing the beliefs of every cutset's value combination, in linear time and space.

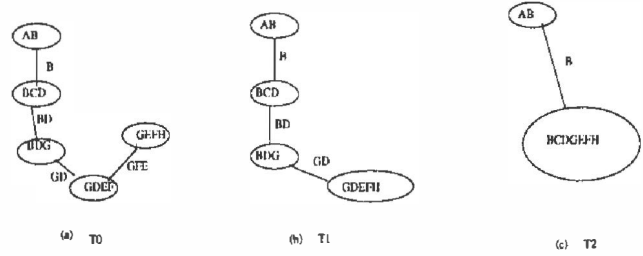

Figure 2: A tree-decomposition with separators equal to (a) 3, (b) 2, and (c) 1

**Lemma 1:** *Given a moral graph and given a cycle-cutset $C$, computing the belief of $C = c$ is time and space linear.*

## 2.3   TRADING SPACE FOR TIME

Assume now that we have a problem whose join-tree has induced width $r$ and separator width $s$ but space restrictions do not allow the necessary $O(exp(s))$ memory required by (modified) tree-clustering. One way to overcome this problem is to collapse cliques joined by large separators into one big cluster. The resulting join-tree has larger subproblems but smaller separators. This yields a sequence of tree-decomposition algorithms parameterized by the sizes of their separators.

**Definition 3**[Primary and secondary join-trees]: Let $T$ be a clique-tree embedding of the moral graph $G$. Let $s_0, s_1, ..., s_n$ be the sizes of the separators in $T$ listed in strictly descending order. With each separator size $s_i$, we associate a tree decomposition $T_i$ generated by combining adjacent clusters whose separator sizes are strictly greater than $s_i$. $T = T_0$ is called the primary join-tree, while $T_i$, when $i > 0$, is a secondary join-tree. We denote by $r_i$ the largest cluster size in $T_i$ minus 1.

Note that as $s_i$ decreases, $r_i$ increases. Clearly, from Theorem 1, it follows that

**Theorem 2:** *Given a join-tree $T$, belief updating and MPE computation can be accomplished using any one of the following time and space bounds $b_i$, where $b_i = (O(n \cdot exp(r_i))$ time, and $O(n \cdot exp(s_i))$ space).* □

We know that finding the smallest tree width of a graph is NP-hard [Arnborg, 1985; Arnborg *et al.*, 1987]; nevertheless, many greedy ordering algorithms provide useful upper bounds that can be inspected in linear time. We denote by $w*_s$ the smallest tree width among all the tree embeddings of $G$ whose separators are of size $s$ or less. Finding $w*_s$ may be hard as well, however. We conclude:

**Corollary 1:** *Given a belief network $BN$, for any $s \leq n$, belief updating and MPE can be computed in time $O(exp(w*_s))$, if $O(exp(s))$ space can be used.* □

Finally, instead of executing a brute-force algorithm to compute the marginal joint probability distributions over the cliques, we can use the conditioning scheme and then record the marginals over the separators only. This leads to the following conclusion:

**Theorem 3:** *Given a constant $s \leq n$, belief assessment and $MPE$ determination can be done in space $O(n \cdot exp(s))$ and in time $O(n \cdot exp(c*_s))$, where $c*_s$ is the maximum cycle-cutset in any subnetwork defined by a cluster in a clique-tree whose separator width is of size $s$ or less, while assuming $s \leq c*_s$.* □

**Example 2:** When Theorem 2 is applied to the belief network in Figure 1 using the join-trees given in Figure 2, we see that finding the beliefs and an MPE can be accomplished in either $O(k^4)$ time and $O(k^3)$ space (based on the primary join-tree $T_0$), or $O(k^5)$ time and $O(k^2)$ space (based on $T_1$), or $O(k^7)$ time and linear space (based on $T_2$). In these cases, the joint distributions over the subproblems defined by the cliques were presumably computed by a brute-force algorithm. If we apply cutset conditioning to each such subnetwork, we get no improvement in the bound for $T_0$ because the largest cutset size in a cluster is 2. For $T_1$, we can improve the time bound from $O(k^5)$ to $O(k^4)$ with only $O(exp(2))$ space (because the cutset size of the subgraph restricted to $\{G, D, E, F, H\}$, is 2); and when applying conditioning to the clusters in $T_2$, we get a time bound of $O(k^5)$ with just linear space (because here, using $T_2$ the cycle-cutset of the whole problem is 3). Thus, the dominating tradeoffs (when considering only the exponents) are between an algorithm based on $T_1$ that requires $O(k^4)$ time and quadratic space and an algorithm based on $T_2$ that requires $O(k^5)$ time and linear space.

shrink The special case of singleton separators was discussed previously in the context of belief networks. When the moral graph can be decomposed to nonseparable components, the conditioning method can be modified to be time exponential in the maximal cutset in each component only [Peot and Shachter, 1991; Darwiche, 1995].

## 2.4  EXAMPLE

We conclude this section by demonstrating in details the mechanics of processing a subnetwork by tree-clustering, by a brute-force methods and by conditioning applied to our example in Figure 1. We will use the join-tree $T - 2$ and process cluster $\{B, C, D, E, F, G, H\}$. Processing this cluster amounts to computing the marginal distribution over the separator $B$, namely (we annotate constant by primes):

$$P(b') = \sum_{d,g,e,c,f,h} (P(g|b',d)P(c|b')P(d|c) \cdot$$
$$P(f|g)P(e|d,f)P(h|g,f,e).$$

Migrating the components as far to the left as possible to exploit a variable elimination scheme which is similar to clustering (for details see [Dechter, 1996]), we get:

$$P(b') = \sum_d \sum_g P(g|b',d) \sum_e \sum_c P(c|b')P(d|c) \cdot$$
$$\sum_f P(f|g)P(e|d,f) \sum_h P(h|g,f,e).$$

**Clustering.** Summing on the variables one by one from right to left, while recording intermediate tables is equivalent to tree-clustering. First, summing over $h$ yields the function $h_H(g,f,e) = \sum_h P(h|g,f,e)$. This takes time exponential in 4 (since there are 4 variables) and space exponential in 3 to record the 3-arity resulting function. Subsequently, summing over $F$ we compute $h_F(g,e,d) = \sum_f P(f|g)P(e|d,f)h_H(g,f,e)$, which is also time exponential in 4 and space exponential in 3. Summing over $C$ yields, $h_C(g,e,d,b') = \sum_c P(c|b')P(d|c)h_F(g,e,d)$ in time exponential in 4 (remember that $b'$ is fixed) and in space exponential in 3. Continuing in this way we compute $h_E(g,d,b') = \sum_e h_C(g,e,d,b')$ (which is time exponential in 3 and space exponential in 2), then, summing over $G$, results in $h_G(d,b') = \sum_g P(g|b',d)h_E(g,d)$ (time exponential in 2 and space exponential in 1) and, finally, $P(b') = \sum_d h_G(d,b')$. Overall, the computation per each variable is time exponential in 4 and space exponential in 3.

**Brute-force.** Alternatively we can do the same computation in a brute-force manner. The same initial expression along the same variable ordering can be used. In this case we expand the probability tree in a forward manner assigning values one by one (conditioning) to all variables, and computing the tuple's probability. We can save computation by using partial common paths of the tree. With this approach we can compute $P(b)$ in time exponential in 6 (the size of the probability tree) and using linear space, since the sums can be accumulated while traversing the tree in a depth-first manner.

**Conditioning.** A third option is to use consitioning within the cluster. Assume we condition on $D, G, E$. This makes the resulting problem a tree. Mechanically, it means that we will expand the probability tree forward using variables $D, G, E$, and for every assignment $g, d, e$ we will use variable elimination (or clustering) in a backwards manner, while treating $d, g, e$ as constant. We perform the backwards computation (denoting constants by primes) as follows. We compute $h_H(g', f, e') = \sum_h P(h|g', f, e')$. This takes time exponential in 2 and space exponential in 1. Subsequently, compute $h_F(g', e', d') = \sum_f P(f|g')P(e'|d', f)h_H(g', f, e')$ which takes time exponential in 1 and constant space. Finally, sum the result over variable $C$: $h_C(g', e', d', b') = \sum_c P(c|b')P(d'|c)h_F(g', e', d')$ in time exponential in 1 and constant space. So far we spent time exponential in 2 and space exponential in 1 at the most. Since we have to repeat this for every value of $g, d, e$ the overall time will be eponential in 5 while the space will be linear.

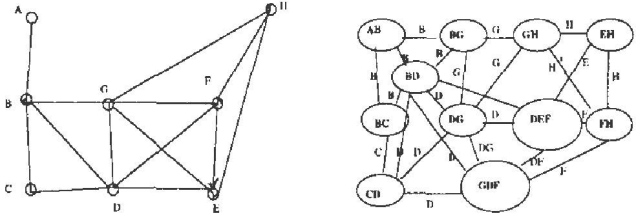

Figure 3: Primal (a) and dual (b) constraint graphs

## 3 CONSTRAINT NETWORKS

**Definition 4** [Constraint network]: A *constraint network* consists of a finite set of variables $X = \{X_1, \ldots, X_n\}$, each associated with a domain of discrete values, $D_1, \ldots, D_n$ and a set of constraints, $\{C_1, \ldots, C_t\}$. A *constraint* is a relation, defined on some subset of variables, whose tuples are all the compatible value assignments. A constraint $C_i$ has two parts: (1) the subset of variables $S_i = \{X_{i_1}, \ldots, X_{i_{j(i)}}\}$, on which the constraint is defined, called a *constraint subset*, and (2) a *relation*, $rel_i$, defined over $S_i : rel_i \subseteq D_{i_1} \times \cdots \times D_{i_{j(i)}}$. The *scheme* of a constraint network is the set of subsets on which constraints are defined. An assignment of a unique domain value to each member of some subset of variables is called an *instantiation*. A consistent instantiation of *all* the variables is called a *solution*. Typical queries associated with constraint networks are to determine whether a solution exists and to find one or all solutions.

**Definition 5** [Constraint graphs]: Two graphical representations of a constraint network are its primal constraint graph and its dual constraint graph. A *primal constraint graph* represents variables by nodes and associates an arc with any two nodes residing in the same constraint. A *dual constraint graph* represents each constraint subset by a node and associates a labeled arc with any two nodes whose constraint subsets share variables. The arcs are labeled by the shared variables.

**Example 3:** Figure 3 depicts the primal and the dual representations of a network having variables $A$, $B$, $C$, $D$, $E$, $F$, $G$, $H$ whose constraints are defined on the subsets $\{(A, B), (B, C), (B, D), (C, D), (D, G), (G, E), (B, G), (D, E, F), (G, D, F), (G, H), (E, H)(F, H)\}$.

Tree-clustering for constraint networks is very similar to tree-clustering for probabilistic networks. In fact, the structuring part is identical. First, the constraint graph is embedded in a clique-tree, then the solution set for each clique is computed. This latter computation is exponential in the clique's size. Therefore, clustering has time and space complexities that are exponential in the induced width of its constraint graph [Pearl, 1988; Lauritzen and Spiegelhalter, 1988].

**Example 4:** Since the graph in Figure 3(a) is identical to the graph in Figure 1(b), it possesses the same clique-tree embeddings. Namely, the maximal cliques of the chordal graph are $\{(A, B), (B, C, D), (B, D, G), (G, D, E, F), (H, G, F, E)\}$ and the join-tree is given in Figure 2(a). The schemes of the subproblems associated with each clique are: $C_{AB} = \{(A, B)\}$, $C_{BCD} = \{(B, C), (C, D)\}$, $C_{BDG} = \{(B, D), (B, G), (G, D)\}$, $C_{GDEF} = \{(G, D)(D, E), (E, F)(G, F)(D, F)\}$ $C_{GEFH} = \{(E, F)(G, F)(G, H)(F, H)(E, H)\}$. As in the probabilistic case, a brute-force application of tree-clustering to the problem is time and space exponential in 4.

The ideas underlying both tree-clustering and conditioning in constraint networks and the implementations of these methods in constraint networks are like those for belief networks. In particular, by refining the clustering method for constraint networks just as we did for probabilistic networks, it is easy to see that clustering in constraint networks obeys similar time and space complexities. Specifically, one way to decide the consistency of a join-tree is to perform directional arc-consistency (also called pair-wise consistency) along some directed rooted tree. If the empty relation is not generated, finding one solution can be done in a backtrack-free manner from root to leaves [Dechter and Pearl, 1987]. The operation of solving each subproblem in the clique-tree and the operation of pair-wise consistency can be interleaved. In this case, constraints may be recorded only on the intersection subsets of neighboring cliques. The time complexity of this modification is exponential in the tree width, while its space complexity is exponentially bounded only by the maximal separator between subproblems. We conclude:

**Theorem 4:** *[Time-space of tree-clustering][Dechter and Pearl, 1989]: Given a constraint problem whose constraint graph can be embedded in a clique-tree having tree width $r$ and separator width $s$, the time complexity of tree-clustering for deciding consistency and for finding one solution is $O(n \cdot exp(r))$ and its space complexity is $O(n \cdot exp(s))$. The time complexity for generating all solutions is $O(n \cdot exp(r) + |solutions|)$, also requiring $O(n \cdot exp(s))$ memory.* $\square$

When the space required by clustering is beyond the available resources, clustering can be coerced to yield smaller separators and larger subproblems, as we have seen earlier, for belief processing. This leads to a conclusion similar to Theorem 2.

**Theorem 5:** *Given a constraint network whose constraint graph can be embedded in a primary clique-tree*

*having separator sizes* $s_0, s_1, ..., s_n$, *whose corresponding maximal clique sizes are* $r_0, r_1, ..., r_n$, *then deciding consistency and finding a solution can be accomplished using any one of the following bounds on the time and space:* $b_i = (O(n \cdot exp(r_i))$ *time,* $O(n \cdot exp(s_i))$ *space).* □

Any linear-space method can replace backtracking for solving each of the subproblems defined by the cliques. One possibility is to use the cycle-cutset scheme. The cycle-cutset method for constraint networks (like in belief networks) enumerates the possible solutions to a set of cutset variables using a backtracking algorithm and, for each consistent cutset assignment, solves a tree-like problem in polynomial time. Thus, the overall time complexity is exponential in the size of the cycle-cutset [Dechter, 1992]. More precisely, the cycle-cutset method is bounded by $O(n \cdot k^{c+2})$, where $c$ is the cutset size, $k$ is the domain size, and $n$ is the number of variables [Dechter, 1990]. Fortunately, enumerating all the cutset's assignments can be accomplished in linear space using backtracking.

**Theorem 6:** *Let $G$ be a constraint graph and let $T$ be a primary join-tree with separator size $s$ or less. Let $c_s$ be the largest minimal cycle-cutset in any subproblem in $T$. Then the problem can be solved in space $O(n \cdot exp(s))$ and in time $O(n \cdot exp(max\{(c_s + 2), s\}))$.*

**Proof:** Since the maximum separator size is $s$, then, from Theorem 4, tree-clustering requires $O(n \cdot exp(s))$ space. Since the cycle-cutset's size in each cluster is bounded by $c_s$, the time complexity is exponentially bounded by $c_s$. However, since time complexity exceeds space complexity, the larger parameter applies, yielding $O(n \cdot exp(max\{(c_s + 2), s\}))$.    □

**Example 5:** Applying the cycle-cutset method to each subproblem in $T_0, T_1, T_2$ shows, as before, that the best alternatives are an algorithm having $O(k^4)$ time and quadratic space, and an algorithm having $O(k^5)$ time but using only linear space.

A special case of Theorem 6, observed before in [Dechter and Pearl, 1987; Freuder, 1985], is when the graph is decomposed into nonseparable components (i.e., when the separator size equals 1).

**Corollary 2:** *If $G$ has a decomposition to nonseparable components in which the size of the maximal cutsets in each component is bounded by $c$, then the problem can be solved in $O(n \cdot exp(c))$ using linear space.* □

## 4  OPTIMIZATION TASKS

Clustering and conditioning are applicable also to optimization tasks defined over probabilistic and deterministic networks. An optimization task is defined relative to a real-valued criterion or cost function associated with every instantiation. In the context of constraint networks, the task is to find a consistent instantiation having maximum cost. In the context of probabilistic networks, the criterion function denotes a utility or a value function, and the task is to find an assignment to a subset of decision variables that maximize the expected criterion function. If the criterion function is decomposable, its structure can be augmented onto the corresponding graph (constraint graph or moral graph) to subsequently be exploited by either tree-clustering or conditioning.

**Definition 6** [Decomposable criterion function [Bacchus and Grove, 1995; D. H. Krantz and Tversky, 1976]]: A criterion function over $n$ variables $X_1, ..., X_n$ having domains of values $D_1, ..., D_n$ is additively decomposable relative to a scheme $Q_1, ..., Q_t$ where $Q_i \subseteq X$ iff

$$f(x) = \sum_{i \in T} f_i(x_{Q_i}),$$

where $T = \{1, ..., t\}$ is a set of indices denoting the subsets of variables, $\{Q_i\}$, $x$ is an instantiation of all the variables. The functions $f_i$ are the components of the criterion function and are specified, in general, by means of stored tables.

**Definition 7** [Constraint optimization, Augmented graph]: Given a constraint network over a set of $n$ variables $X = X_1, ..., X_n$ defined by a set of constraints $C_1, ..., C_t$ having scheme $S_1, ..., S_n$, and given a criterion function $f$ decomposable over $Q_1, ..., Q_t$, the constraint optimization problem is to find a consistent assignment $x$ such that the criterion function is maximized. The *augmented constraint graph* contains a node for each variable and an arc connects any two variables that appear either in the same constraint component $S_i$ or in the same function component $Q_i$.

Since constraint optimization can be performed in linear time when the augmented constraint graph is a tree, both tree-clustering and conditioning can extend the method to non-tree structures [Dechter *et al.*, 1990]. We can conclude:

**Theorem 7:** [Time-space of constraint optimization]: *Given a constraint optimization problem whose augmented constraint graph can be embedded in a clique-tree having tree width $r$ and separator width $s$ and a cycle-cutset size $c$, the time complexity of finding an optimal consistent solution using tree-clustering is* $O(n \cdot exp(r))$ *and space complexity* $O(n \cdot exp(s))$. *The time complexity for finding a consistent optimal solution using conditioning is* $O(n \cdot exp(c))$ *while its space complexity is linear.* □

In a similar manner, the structure of the criterion function can augment the moral graph when computing the maximum expected utility (MEU) of some decisions in a general influence diagram [Shachter, 1986]. An *influence diagram* is a belief network having decision variables as well as an additively decomposable utility function.

**Definition 8** [Finding the MEU] Given a belief network $BN$, and a real-valued utility function $u(x)$ that is additively decomposable relative to $Q_1, ..., Q_t$, $Q_i \subseteq X$, and given a subset of decision variables $D = \{D_1, ...D_k\}$ that are root variables in the directed acyclic graph of $BN$, the MEU task is to find an assignment $d^o = (d^o{}_1, ..., d^o{}_k)$ such that

$$(d^o) = argmax_{\bar{d}_k} \sum_{x_{k+1}, ..., x_n} \Pi_{i=1}^k P(x_i|x_{pa(X_i)}, \bar{d}_k)u(x).$$

The utility-augmented graph of an influence diagram is its moral graph with some additional edges; any two nodes appearing in the same component of the utility function are connected as well.

A linear-time propagation algorithm exists for the MEU task whenever the utility-augmented moral graph of the network is a tree [Jennsen and Jennsen, 1994]. Consequently, by exploiting the augmented moral graph, we can extend this propagation algorithm to general influence diagrams. The two approaches that extend this propagation algorithm to multiply-connected networks, cycle-cutset and conditioning, are applicable here as well [Pearl, 1988; Lauritzen and Spiegelhalter, 1988; Shachter, 1986]. It has also been shown that elimination algorithms are similar to tree-clustering methods [Dechter and Pearl, 1989]. In summary:

**Theorem 8:** [Time-space of finding the MEU]: *Given a belief network having a subset of decision variables, and given an additively decomposable utility function whose augmented moral graph can be embedded in a clique-tree having tree width $r$ and separator width $s$ and a cycle-cutset size $c$, the time complexity of computing the $MEU$ using tree-clustering is $O(n \cdot exp(r))$ and the space complexity is $O(n \cdot exp(s))$. The time complexity for finding a $MEU$ using conditioning is $O(n \cdot exp(c))$ while the space complexity is linear.* □

Once we have established the graph that guides clustering and conditioning for either constraint optimization or finding the MEU, the same principle of trading space for time becomes applicable and will yield a collection of algorithms governed by the primary and secondary clique-trees and cycle-cutsets of the augmented graphs as we have seen before.

The following theorem summarizes the time and space tradeoffs associated with optimization tasks.

**Theorem 9:** *Given a constraint network (resp., a belief network) and given an additively decomposable criterion function $f$, if the augmented constraint graph (resp., moral graph) relative to the criterion function can be embedded in a clique-tree having separator sizes $s_0, s_1$, and corresponding maximal clique sizes $r_0, r_1, ..., r_n$ and corresponding maximal minimal cutset sizes $c_0, c_1, ..., c_n$, then finding an optimal solution (resp., finding the maximum expected criterion value) can be accomplished using any one of the following*

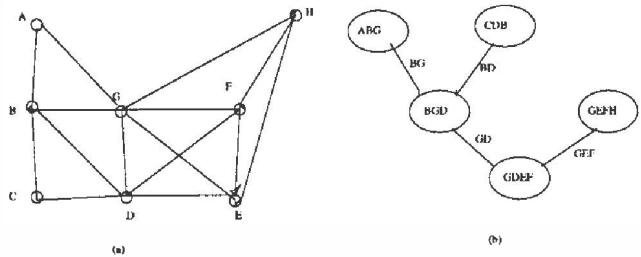

Figure 4: An augmented moral graph for the utility function $f(a, b, c, d, e, f, g.h) = a \cdot g + c^2 + 5d \cdot e \cdot f$

*bounds on the time and space: if a brute-force approach is used for processing each subproblem the bounds are $b_i = (O(n \cdot exp(r_i))$ time, $O(n \cdot exp(s_i))$ space); if conditioning is used for each cluster, the bounds are $b_i = (O(n \cdot exp(c_i))$ time, $O(n \cdot exp(s_i))$ space).* □

**Example 6:** If we define a criterion function whose components are singleton variables, then the complexity bounds of optimization are identical to constraint satisfaction or finding the posterior probabilities. If we have the following criterion function defined on the belief network in Figure 1

$$f(a, b, c, d, e, f, g, h) = a \cdot g + c^2 + 5d \cdot e \cdot f$$

then the augmented moral graph will have one additional edge connecting nodes $A$ and $G$ (see Figure 4(a)), resulting in a primary clique-tree embedding in Figure 4(b) that differs from the tree in Figure 2(a).

## 5   CONCLUSIONS

We have shown that both constraint network processing and belief network processing obey a time-space tradeoff based on structural properties that allow tailoring a combination of tree-clustering and cycle-cutset conditioning to certain time and space requirements. The same kind of tradeoff is obeyed by optimization problems when using the problem's graph augmented with arcs reflecting the structure of the criterion function.

Various algorithms that combine tree-clustering with conditioning were proposed in the past in the context of constraint networks [Jegou, 1990] and belief networks [Darwiche, 1995; Peot and Shachter, 1991]. These algorithms are normally space linear (a point that is not always appreciated), and they seem to fall in the first tradeoff class, as they exploit singleton separators only. Our analysis presents a spectrum of algorithms that will allow a richer time-space performance balance.

We would like to address breifly two questions. The first question is to what extent real-life problems possess structural properties that can be exploited by either clustering, conditioning, or their hybrids. Recent empirical investigation using the domain of circuit diagnosisconfirm that the domain of circuit analysis can benefit substantially from structure-based algorithms. The results are presented in a companion paper [Fattah and Dechter, 1996].

The second question is to what extent do worst-case bounds indeed reflect on average-case performance. Previous experimental work with clustering and conditioning shows that while average-case performance of clustering methods correlates well with worst-case performance, conditioning methods are sometimes much more effective than predicted by worst-case analysis. It is therefore necessary to implement and test the algorithms implied by our analysis.

### Acknowledgment

I would like to thank Rachel Ben-Eliyahu and Irina Rish for useful comments on the previous version of this paper. This work was partially supported by NSF grant IRI-9157636, Air Force Office of Scientific Research grant, AFOSR F49620-96-1-0224, and Rockwell MICRO grant #ACU-20755 and 95-043.